\documentclass[twoside,11pt]{article}

\usepackage[preprint]{jmlr2e}

\usepackage{graphicx}
\usepackage{adjustbox}
\usepackage{float}

\ShortHeadings{Graph-Attentive MAPPO for Dynamic Retail Pricing}{Krishna Kumar Neelakanta}
\firstpageno{1}

\begin{document}

\title{Graph-Attentive MAPPO for Dynamic Retail Pricing}

\author{\name Krishna Kumar Neelakanta Pillai Santha Kumari Amma \\
       \email krishnakumarns82@gmail.com \\
       \addr Independent Researcher\\
       19540 Highland Oaks Drive, Apt 202\\
       Estero, Florida, 33928, USA}

\maketitle

\begin{abstract}
Dynamic pricing in retail requires policies that adapt to shifting demand while coordinating decisions across related products. We present a systematic empirical study of multi-agent reinforcement learning for retail price optimization, comparing a strong MAPPO baseline with a graph-attention-augmented variant (MAPPO+GAT) that leverages learned interactions among products. Using a simulated pricing environment derived from real transaction data, we evaluate profit, stability across random seeds, fairness across products, and training efficiency under a standardized evaluation protocol. The results indicate that MAPPO provides a robust and reproducible foundation for portfolio-level price control, and that MAPPO+GAT further enhances performance by sharing information over the product graph without inducing excessive price volatility. These results indicate that graph-integrated MARL provides a more scalable and stable solution than independent learners for dynamic retail pricing, offering practical advantages in multi-product decision-making.
\end{abstract}


\section{Introduction}
In retail, dynamic pricing aims to modify product prices in response to changing operational limitations, portfolio dynamics, and demand. A merchant who oversees dozens or hundreds of SKUs must coordinate pricing decisions across related items whose sales collectively react to seasonality, promotions, and latent cross-effects (such as complements and replacements), in contrast to single-product settings. Creating stable and profitable policies that can be replicated under actual data settings is still a practical issue.
   
\indent Multi-agent reinforcement learning (MARL) offers a natural framework for portfolio-level price control: each product can be treated as an agent with local observations and a shared objective. Among MARL baselines, Multi-Agent Proximal Policy Optimization (MAPPO) is a strong and widely used method due to its stable updates and centralized-training/decentralized-execution design. However, vanilla MARL often treats agents independently at the function-approximation level, limiting the ability to exploit inter-product structure that is known to drive retail outcomes.
   
    To address this gap, we study a graph-aware MARL approach for retail pricing that augments MAPPO with graph attention inside the policy and value networks (MAPPO+GAT). Intuitively, graph attention allows each product-agent to condition its pricing decision on learned representations of related products, enabling information sharing about demand trends, price sensitivities, and seasonal dynamics.
     
    \indent Our evaluation is based on a simulated pricing scenario that was created using actual transaction data. We use common supervised diagnostics to confirm that the dataset shows learnable demand structure prior to policy training. After that, we compare MAPPO and MAPPO+GAT using a consistent assessment technique, reporting not just the average test profit but also price-stability metrics that punish excessive oscillations, robustness across random seeds, and fairness between products (such as the balance of profit/volume). Retailer values are reflected in this wider lens: reliable client experiences and equitable results across the catalog are just as important as profit.
     
    \indent Our contributions are threefold:
\begin{itemize}
\item Methodology: A simple, reproducible graph-attention augmentation of MAPPO for dynamic pricing in retail, enabling cross-product signal sharing inside the policy/value networks.
\item Evaluation: A variance-aware empirical protocol on a real-data-based environment, with metrics spanning profit, fairness, and stability in addition to standard learning curves.
\item Evidence \& Assets: An empirical study showing consistent improvements of graph-aware MARL over a strong MAPPO baseline, plus open scripts for diagnostics, paired statistical analysis, fairness, and stability.
\end{itemize}

\indent In sum, we show that the pricing policies which encode cross- product structure can deliver meaningful, practice-relevant gains without sacrificing stability or reproducibility. By embedding graph attention within a strong MARL baseline, our study demonstrates that portfolio-level price control benefits from learned interactions among products, not just per-item features.

\section{Literature Review}
\subsection{Dynamic Pricing with Reinforcement Learning}
Reinforcement learning (RL) has been used for price control in situations with random demand and changes over time. The research on demand response showed that RL can effectively assist in designing tariffs and engaging customers. This provided initial proof that learning-based pricing can manage uncertainty and operational limits ([\cite{1,2}]). Although these studies focus on different aspects than retail SKU pricing, they highlight RL as a solid method for price control.

\subsection{Multi-Agent RL (MARL) for Co-ordinated Pricing}
Multi-agent reinforcement learning (MARL) deals with non-stationarity and coordination by employing centralized training with decentralized execution (CTDE). Foundational surveys outline the algorithmic challenges and stability issues in MARL ([ \cite{3}]). Among CTDE algorithms, Proximal Policy Optimization (PPO) ([ \cite{4}]), known as MAPPO, has shown to be a strong and stable baseline for cooperative tasks. It reduces the instability of independent learners while maintaining scalability ([ \cite{5}]). Researchers have also looked into attention mechanisms to help with selective information sharing and credit assignment among agents ([ \cite{6}]). However, most MARL implementations for pricing do not clearly represent product-level relationships within the policy and value networks.

\subsection{Graph Modeling of Product Relationships.}
Retail portfolios exhibit cross-product effects (substitution/complementarity, promotional spillovers). Graph neural networks (GNNs) provide an inductive bias for such structure, propagating information over item relations; many surveys document their representational advantages for relational data ([ \cite{7}]). Classic GCN aggregates neighbors with fixed weights ([ \cite{9}]), GraphSAGE introduces sampled aggregation for inductive settings ([ \cite{10}]), and GIN improves discriminative power via sum aggregation ([ \cite{11}]). These models, however, use uniform or fixed neighbor influence and are primarily applied to prediction (e.g., recommendation, demand forecasting), rather than control via pricing policies.

\subsection{Why Graph Attention for Retail Pricing}
Graph Attention Networks (GAT) learn data-driven, context-dependent weights over a node's neighbors, enabling each product to emphasize the most relevant related items at a given time (e.g., seasonality, promotions, transient cross-elasticities) ([ \cite{8}]). This selective aggregation is well-suited to retail pricing, where the importance of substitutes and complements is state-dependent. Compared with GCN/GraphSAGE/GIN, GAT provides a lightweight, parameter-efficient mechanism to adapt neighbor influence during policy evaluation, which is desirable when embedding relational information directly into control.

\subsection{Positioning and Gap}
Taken together: (i) RL is a credible approach to pricing under uncertainty (\cite{1,2}); (ii) MAPPO offers a robust MARL baseline for coordinated control (\cite{5}); and (iii) GNNs, especially GAT, provide principled means to model item relations ([ \cite{7,8}]). What remains comparatively underexplored is a graph-aware MARL framework that embeds GAT inside MAPPO so each SKU's policy/value computation can attend to learned product interactions—moving beyond per-item features and fixed neighbor aggregation. Our work addresses this gap by evaluating MAPPO+GAT architecture for dynamic pricing in retail, with an empirical protocol that reports profit and practitioner-oriented metrics (fairness, price stability) on a real-data-derived environment, and with configurations suitable for modest catalogs.

\section{Problem Formulation}

\subsection{Setting and Notation}
We consider a single retailer managing a portfolio of $n$ products (SKUs). Let the index set be V={1,…,n}. Time is discrete, t=1,…,H, where H is the episode horizon (e.g., a rolling window over calendar days).
\begin{itemize}
    \item $p_{i,t} \in \mathbb{R}_+$: price of SKU i at time t.
    \item $q_{i,t} \in \mathbb{R}_+$: realized demand (units sold) of SKU i at time t.
    \item $c_i \in \mathbb{R}_+$: unit cost of SKU i(assumed known and time-invariant).
    \item $x_{i,t}$: exogenous features (e.g., seasonality indicators, temporal trends, historical sales signals).
    \item $N(i)$: neighbor set of SKU i in an item graph $G=(V,E)$ capturing cross-product relations (e.g., co-purchase).
    \item $\gamma \in (0,1]$: discount factor (typically $\gamma=1$ for finite-horizon profit).
\end{itemize}

Action space: Each SKU i chooses a price from a finite set of multipliers around a reference level $p_{\bar{i}}$:

    \[A_i = \{\alpha_k p_{\bar{i}} : \alpha_k \in M\}, M \subset \mathbb{R}_+\]

The joint action is \[a_t = (p_{1,t}, \ldots, p_{n,t}) \in A_1 \times \cdots \times A_n\]

State and observations. The environment state $s_t$ aggregates all SKUs’ features and latent drivers; SKU i observes $o_{i,t}$, a local view constructed from $x_{i,t}$, recent lags/EMAs, and (optionally) graph-aggregated neighbor summaries.

\subsection{Retail Dynamics and Demand Oracle}
We assume sufficient stock (no stockouts) and a single-retailer market (competitor behavior, if any, is absorbed into $x_{i,t}$). The per-SKU demand is generated by a data-driven oracle fitted on historical transactions (e.g., CatBoost), yielding the stochastic quantity:

\[q_{i,t} = f_i (x_{i,t}, p_{i,t}, \{p_{j,t} \}_{j \in N(i)}) + \epsilon_{i,t}\]

where $f_i(\cdot)$ captures own-price and cross-SKU effects and $\epsilon_{i,t}$ is zero-mean noise. This defines the transition kernel $P(s_{t+1}|s_t,a_t)$.

\subsection{Graph-Structured Agents}

The SKU relations are encoded by a fixed, sparse graph $G=(V,E)$. Let $h_{i,t}$ be a local embedding built from $o_{i,t}$. A graph attention operator $\Phi$ produces a neighbor-aware representation

\[z_{i,t} = \Phi(h_{i,t}, \{h_{j,t} \}_{j \in N(i);G})\]

where $\Phi$ computes data-dependent attention weights over N(i), allowing SKU i to emphasize contextually relevant neighbors (e.g., substitutes during promotions). The observation supplied to control is $\tilde{o}_{i,t}=[o_{i,t};z_{i,t}]$.

\subsection{Objective and Constraints}

The per-period profit is

\[r_t = \sum_{i=1}^n (p_{i,t} - c_i) q_{i,t} - \lambda_{stab} \sum_{i=1}^n \left| \frac{p_{i,t} - p_{i,t-1}}{p_{i,t-1}} \right|\]

with optional $\lambda_{\mathrm{stab}} \geq 0$ to discourage excessive price oscillations. Prices can be constrained by bounds $\underline{p}_i \leq p_{i,t} \leq \overline{p}_i$ (implicitly enforced via $A_i$).

The retailer seeks a decentralized joint policy $\pi = \{\pi_i\}_{i=1}^n$, where $\pi_i(a_{i,t}|\tilde{o}_{i,t})$, that maximizes expected discounted profit:

\[ \max_{\pi} J(\pi) = E_{s_1 \sim \rho, a_{1:H} \sim \pi, s_{2:H} \sim P} \left[ \sum_{t=1}^H \gamma^{t-1} r_t \right] \]

\subsection{Learning Problem (CTDE / MAPPO)}

We adopt centralized training with decentralized execution (CTDE). During training, a centralized value function $V_\psi(s_t)$ (or critic $V_\psi(s_t,a_t)$) conditions on global information to stabilize updates; at execution, each agent uses only $\tilde{o}_{i,t}$. Concretely, we employ Multi-Agent PPO (MAPPO) to optimize $\pi = \{\pi_i\}$ with clipped policy updates and a shared/centralized critic. The graph attention module $\Phi$ is embedded inside the actor/critic feature pipelines so that policies and value estimates are graph-aware but remain decentralized at test time.

\subsection{Assumptions}

\begin{itemize}
    \item Single retailer: competitor dynamics are not explicitly modeled.
    \item No inventory coupling: stock is ample over the horizon (no lost sales).
    \item Static graph $G$: relations reflect co-behavior in the historical window and are fixed during training/evaluation.
    \item Discrete price grid: actionable price multipliers $M$ encode practical bounds and step sizes.
    \item Finite-horizon episodes: evaluation rolls over a fixed test window with randomized starts.
\end{itemize}

This formulation captures portfolio-level price control with cross-SKU interactions via a graph-aware observation. The GAT component enables context-dependent neighbor influence, while MAPPO provides stability and scalability in the multi-agent optimization.

\section{Methodology}
We develop a graph-aware multi-agent control framework for dynamic pricing in retail. Each product (SKU) is an agent; policies are trained with Multi-Agent Proximal Policy Optimization (MAPPO) under centralized training and decentralized execution (CTDE). To capture cross-product interactions, we embed a Graph Attention Network (GAT) layer inside the actor-critic feature stacks, enabling context-dependent aggregation of neighbor information on an item graph derived from transactions.

\subsection{Dataset Trimming and Sanity Checks}
We construct a self-contained slice of Online Retail II spanning 2011-07-01 to 2011-12-09 and retain the top 60 SKUs by activity. After cleaning (removing cancelled/negative lines and missing IDs), the trimmed dataset contains 38,794 transactions, 7,903 invoices, and 2,949 customers. The item graph built from same-invoice co-occurrences yields 720 directed edges with top-k = 12 per SKU; all 60 SKUs lie in the largest weakly connected component. 

Before RL training we fit per-SKU CatBoost demand models on a chronological split (test window 2011-11-21 to 2011-12-09). On the held-out test set we obtain R² (log1p) = 0.8165, RMSE = 131.39 (qty), MAPE = 1.03\%; quantity-weighted RMSE/ MAPE are 206.34 and 0.87\%, indicating strong learnability and non-trivial seasonality.

\subsection{Item Graph Construction}
Let $G=(V,E)$  with $|V| = n$ SKUs. We form edges from co-purchase statistics at the invoice level:
\begin{itemize}
    \item Weight $w_{ij}$: co-occurrence count or normalized lift.
    \item Sparsification: retain edges with $w_{ij} \geq \tau$, then top-k outgoing neighbors per node (small k for tractability).
    \item Direction: directed edges after top-k pruning; $G$ is held static during training/evaluation.
    \item Co-purchase captures complements/substitutes and shared demand shocks; top-k preserves salient ties and bounds compute.
\end{itemize}

\subsection{Observation and Action Design}
Observation $o_{i,t}$:  per-SKU vector including:
\begin{itemize}
    \item price features (log price, normalized price vs. median),
    \item demand history (lags, EMA, trend),
    \item calendar/seasonal context (DoW/month encodings), and
    \item optional neighbor summaries (pre-GAT statistics for stability).
\end{itemize}

\textbf{Action $a_{i,t}$:} discrete price multipliers from a bounded set $M$ applied to a SKU-specific reference price $\bar{p}_i$. This encodes practical step sizes and legal/operational bounds.

\textbf{Reward:} profit at epoch $t$: $r_t = \sum_{i=1}^n (p_{i,t} - c_i) q_{i,t} - \lambda_{\mathrm{stab}} \sum_{i=1}^n | \Delta \log p_{i,t} |$, with optional stability penalty $\lambda_{\mathrm{stab}} \geq 0$.

\subsection{Demand Oracle (Environment Dynamics)}
We fit a per-SKU supervised demand model $f_i(\cdot)$ on historical data to map $(o_{i,t},a_{i,t},\{a_{j,t}\}_{j \in N(i)})$ to $q_{i,t}$. We use gradient-boosted trees (CatBoost) with log1p(quantity) targets and chronological splits. The oracle provides stochastic transitions $P(s_{t+1}|s_t,a_t)$ by sampling noise around $\hat{q}_{i,t}=f_i(\cdot)$. This preserves empirical seasonality and cross-SKU effects and yields a stationary, reproducible simulator for policy learning.

\subsection{Baseline: MAPPO (CTDE)}

 We adopt MAPPO with:
 \begin{itemize}
    \item Actor: per-agent policy $\pi_i(a_{i,t}|\tilde{o}_{i,t})$.
    \item Critic: shared centralized value $V_\psi(s_t)$ (or centralized state-action value), using global or concatenated features to stabilize training.
    \item Optimization: PPO clipped objective with GAE($\lambda$), entropy regularization, and minibatch SGD over rollouts; parallel environments for throughput.
    \item Execution: decentralized (each agent uses only its local, graph-augmented observation).
\end{itemize}

\subsection{MAPPO+GAT Architecture}

 \begin{figure}[htbp]
    \centering
    \includegraphics[height=0.4\textheight,keepaspectratio]{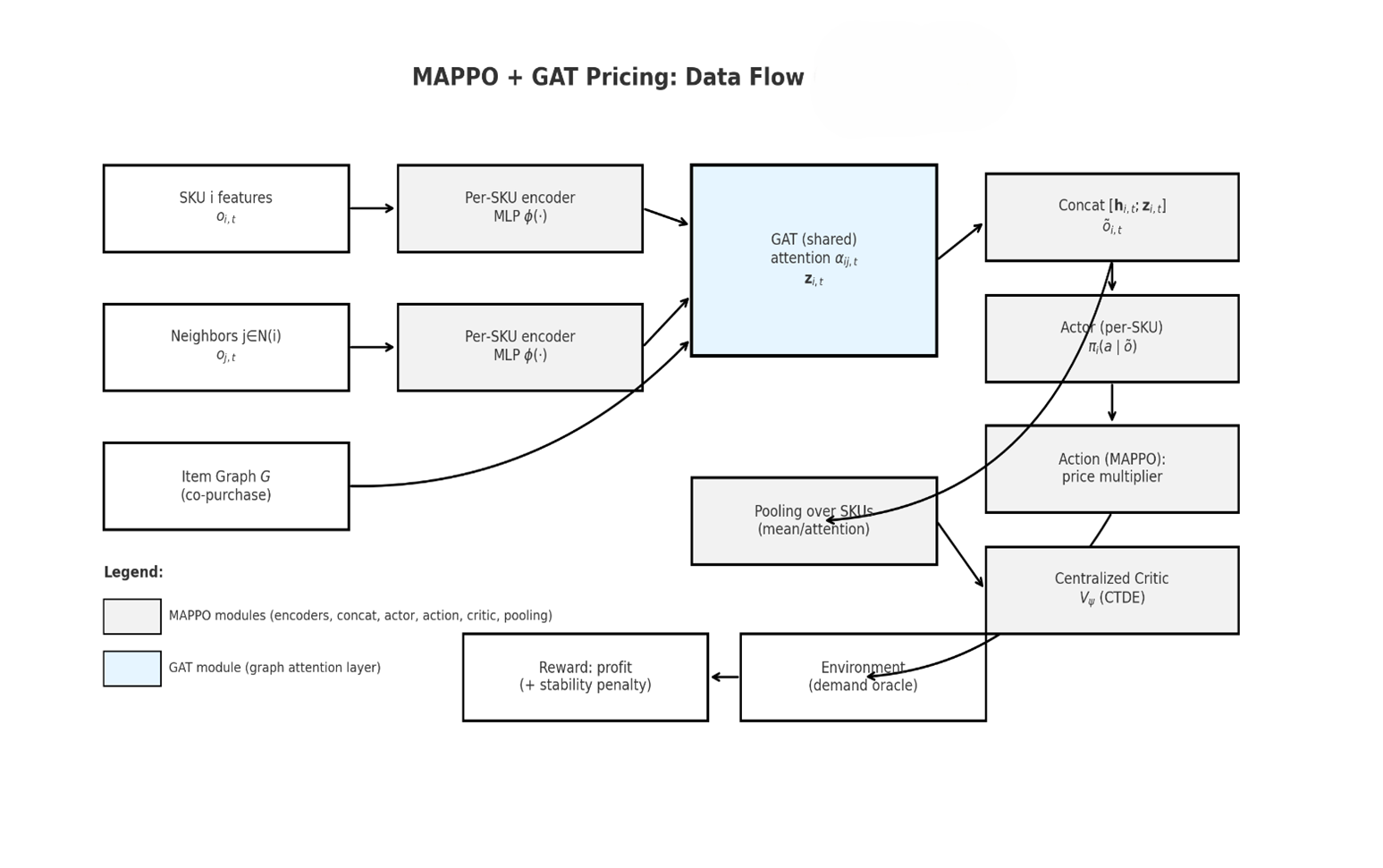}
    \label{fig:fig_0}
    \end{figure}

The architecture integrates graph attention with a centralized-training, decentralized-execution (CTDE) actor-critic. Each SKU is an agent with a local observation; a shared Graph Attention Network (GAT) layer provides context-dependent aggregation of neighbor information on the item graph. The objective is to exploit cross-product structure while retaining the stability of a strong MAPPO baseline.
    
\textbf{Feature stacks:} For each SKU $i$ at time $t$, we compute a local embedding $h_{i,t}=\phi(o_{i,t})$ via a small MLP $\phi(\cdot)$. A single, shared GAT block produces a neighbor-aware representation
   
    \[z_{i,t} = \mathrm{Concat}_{h=1}^H \left( \sum_{j \in N(i)} \alpha_{ij,t}^{(h)} W^{(h)} h_{j,t} \right) \quad \alpha_{ij,t}^{(h)} = \mathrm{softmax}_j \left( \mathrm{LeakyReLU}(a^{(h)\top} [W^{(h)} h_{i,t} \Vert W^{(h)} h_{j,t}]) \right)\]
    
where H is the number of heads, N(i)is the pruned neighbor set (top-k edges), and $W^{(h)}, a^{(h)}$ are head-specific parameters. The actor input is $\tilde{o}_{i,t}=[h_{i,t};z_{i,t}]$.
    
\textbf{Actor and action parameterization:} The policy $\pi_{\theta,i}(a_{i,t}|\tilde{o}_{i,t})$ outputs a categorical distribution over discrete price multipliers $M$ around a SKU-specific reference price. We use a lightweight MLP head on top of $\tilde{o}_{i,t}$ with entropy regularization to discourage premature collapse.
    
\textbf{Centralized critic:} A shared critic $V_\psi(s_t)$ stabilizes training by conditioning on a global summary. We consider two practical variants:
\begin{itemize}
\item Pooled graph features: mean- or attention-pool the set $\{\tilde{o}_{i,t}\}_{i=1}^n$ to form a permutation-invariant global vector;
\item Concatenated summary: project each $\tilde{o}_{i,t}$ then aggregate via mean/max pooling.
\end{itemize}
The critic MLP regresses Monte-Carlo returns with GAE($\lambda$) advantages.

\textbf{Parameter sharing and regularization:}
\begin{itemize}
    \item Sharing: GAT parameters are shared across agents; optionally shared between actor and critic to reduce memory.
    \item Sparsity \& stability: top-k neighbors, dropout on attention coefficients, $L_2$ weight decay on GAT projections, and edge-drop during training.
    \item Capacity: one GAT layer; hidden size $d \in \{32, 64\}$; heads $H \in \{2, 4\}$. This keeps compute practical for modest catalogs.
\end{itemize}
    
\textbf{Training objective (MAPPO):} We optimize the clipped PPO objective with minibatch SGD over parallel rollouts:

    \[L_{PPO}(\theta) = E \left[ \min \left( r_t(\theta) A_t, \mathrm{clip}(r_t(\theta), 1-\epsilon, 1+\epsilon) A_t \right) \right] - c_v L_V(\psi) + c_e H[\pi_\theta]\]
    
    where $r_t(\theta) = \frac{\pi_\theta(a_t|\tilde{o}_t)}{\pi_{\theta_{\mathrm{old}}}(a_t|\tilde{o}_t)}$, $A_t$ are GAE advantages from the centralized critic, and $H$ is entropy. Execution is fully decentralized (each agent uses only $\tilde{o}_{i,t}$).
    
\textbf{Complexity:} For $n$ SKUs, top-k neighbors, hidden $d$, and $H$ heads, the GAT block adds $O(nkHd)$ messages per step. With one layer, small $d$, $H$, and sparse $k$, overhead is modest relative to MAPPO MLPs—appropriate for small retailers.
    
\indent GAT is preferred over GCN/GraphSAGE/GIN because attention yields context-dependent neighbor weights, matching retail scenarios where the relevance of substitutes/complements varies with season and current prices.
    
\subsection{Training Protocol and Hyperparameters}
\begin{itemize}
    \item Rollouts: $T$ steps per update with $N$ parallel environments; advantage estimation via GAE.
    \item Updates: PPO clipping $\epsilon \in [0.1, 0.2]$, value loss coefficient $c_v$, entropy coefficient $c_e$.
    \item Batching: Minibatch size tuned to fit a single consumer-grade GPU/CPU; mixed precision optional.
    \item GAT capacity: hidden size $d \in \{32, 64\}$, heads $H \in \{2, 4\}$, top-k $\in [8, 12]$ neighbors; one GAT layer by default.
    \item Early stopping / model selection: best validation profit checkpoint per seed.
    \item Seeds: multiple seeds for robustness; results reported as mean/median with confidence intervals.
\end{itemize}

\subsection{Evaluation Protocol}
\begin{itemize}
    \item Test Horizon -We evaluate on a fixed calendar window corresponding to the held-out period. Episodes are generated by rolling start indices within this window to obtain multiple, non-overlapping trajectories.
    \item Variance Control-To reduce estimator variance and enable paired comparisons, we employ Common Random Numbers (CRN): for each seed, MAPPO and MAPPO+GAT are evaluated under identical episode initializations and RNG streams.
    \item Primary Metric -The principal outcome is cumulative test profit aggregated over each episode and averaged across episodes/seeds.
    \item Robustness Reporting - We report on the median and mean test profit, the seed-wise win rate of MAPPO+GAT over MAPPO, and 95\% confidence intervals for the paired difference (MAPPO+GAT $-$ MAPPO) under CRN.
    \item Practice-Oriented Indicators - We quantify fairness across products via Jain’s index computed on per-SKU profits, and price stability via the mean absolute percentage price change between consecutive decisions. We additionally include an ablation with a volatility penalty in the reward to illustrate the profit–stability trade-off.
\end{itemize}

\section{Experimental Setup}

\subsection{Data and Simulator}
We derive a single-retailer pricing simulator from the Online Retail II dataset. After cleaning cancelled/negative lines and missing IDs, we trim to a Jul 1–Dec 9, 2011 window with the top-60 SKUs by activity (38,794 transactions; 7,903 invoices; 2,949 customers). A per-SKU CatBoost demand model (chronological split) serves as the stochastic demand oracle, preserving seasonality and cross-SKU effects. The test window is Nov 21–Dec 9, 2011.

\subsection{Item Graph}
We construct a fixed, directed co-purchase graph at the invoice level; edges are retained for co-occurrence counts $\geq \tau$ and then pruned to top-k = 12 outgoing neighbors per SKU. All 60 SKUs belong to the largest weakly connected component.

\subsection{Observations and Actions}
Each agent (SKU) observes price features (log/normalized price), recent sales statistics (lags, EMA, trend), and calendar encodings (DoW, month). The action space is a discrete set of price multipliers about a SKU-specific reference price $\bar{p}_i$(bounded within operational limits). Prices remain within retailer-plausible ranges.

\subsection{Reward}
Per-period reward is profit $\sum_{i=1}^n (p_i - c_i) q_i$ with an optional price-stability 
penalty $\lambda_{\mathrm{stab}} \sum_{i=1}^n | \Delta \log p_i |$.

\subsection{Baselines and Proposed Method}
\begin{itemize}
    \item MAPPO (CTDE): per-SKU actors with a centralized value function.
    \item MAPPO+GAT (proposed): MAPPO with a single shared Graph Attention layer embedded in the feature stack (actor and critic), using sparse top-k neighbors and multi-head attention. Architectures are capacity-matched (lightweight hidden sizes and heads) for fairness and workstation feasibility.
\end{itemize}

\subsection{Training Protocol}
We employ PPO with clipped updates under CTDE: generalized advantage estimation (GAE), entropy regularization, and minibatch SGD. Unless noted, we train 15 random seeds with 60k environment steps per seed. Rollouts, epochs per update, and minibatch sizes follow the configuration tuned for a home workstation. Model selection uses the best validation profit checkpoint per seed.

\subsection{Evaluation Protocol}
We evaluate on the held-out calendar window with rolled episode starts. To reduce variance and enable paired statistics, we use Common Random Numbers (CRN): for each training seed, MAPPO and MAPPO+GAT are evaluated on the same set of episode initializations and RNG streams. We run 50–100 episodes per seed under CRN and report episode-averaged metrics. Primary metric is cumulative test profit; robustness metrics include median, seed-wise win rate, and 95\% CIs for paired differences (GAT $-$ MAPPO). Practice-oriented metrics include Jain’s index over per-SKU profits (fairness) and mean absolute \% price change (stability). Volatility-penalty ablations sweep $\lambda_{\mathrm{stab}}$ to trace the profit–stability frontier.

\subsection{Implementation and Hardware}
Experiments are implemented in Python with PyTorch (RL) and CatBoost (supervised oracle). Runs are executed on a home workstation (Windows/Anaconda; CPU). Sparse graph ops (top-k) and a single GAT layer keep overhead modest for tens of SKUs.

\section{Results and Observations}
We compare MAPPO and MAPPO+GAT on a retail environment derived from UCI Online Retail II (60 SKUs). Models are selected by best validation performance at 60k steps and evaluated on the held-out test window using Common Random Numbers (CRN): 100 test episodes per seed, paired across methods; 15 seeds in total. Primary metric is cumulative test profit. We also report fairness (Jain's index across SKU profits) and price stability (mean absolute \% price change), both averaged per seed.

\begin{figure}[htbp]
\centering
\includegraphics[height=0.4\textheight,keepaspectratio]{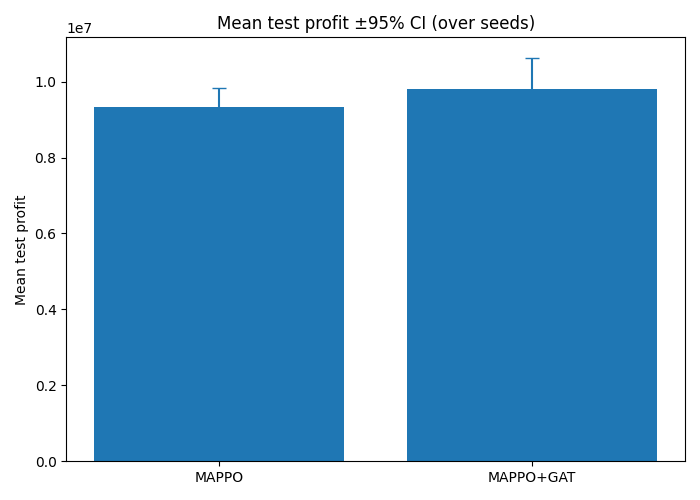}
\caption{Mean test profit ($\pm$95\% CI) across seeds for MAPPO and MAPPO+GAT.}
\label{fig:fig_1}
\end{figure}

\begin{figure}[htbp]
\centering
\includegraphics[height=0.4\textheight,keepaspectratio]{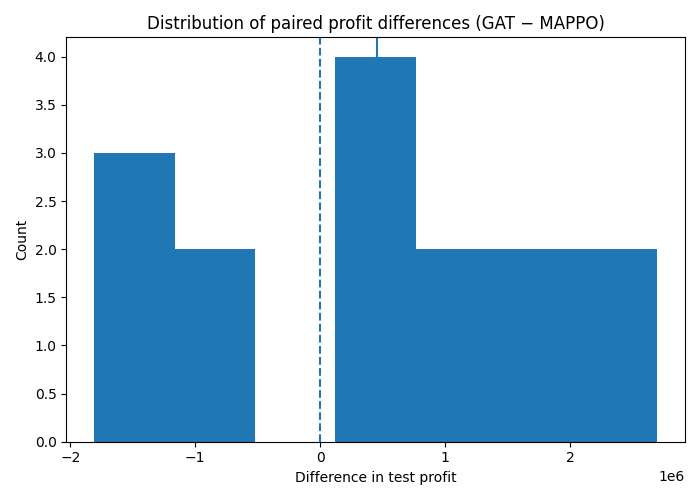}
\caption{Histogram of paired profit differences (GAT $-$ MAPPO) over seeds (CRN-paired episodes).}
\label{fig:fig_2}
\end{figure}

\begin{figure}[htbp]
\centering
\includegraphics[height=0.4\textheight,keepaspectratio]{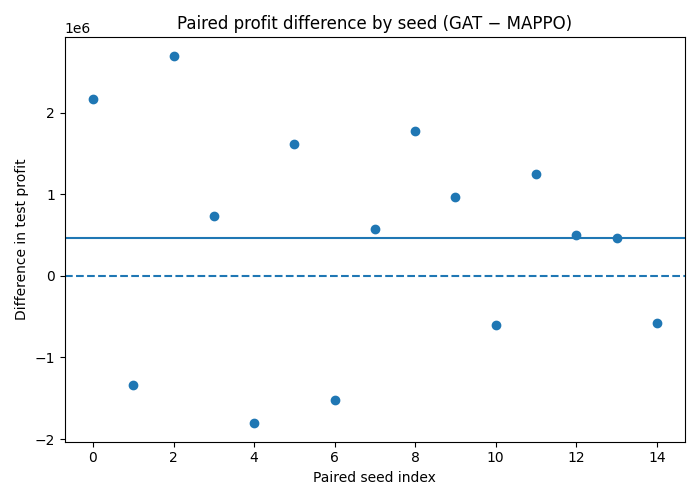}
\caption{Per-seed paired difference (GAT $-$ MAPPO) in test profit; solid line = mean, dashed = zero.}
\label{fig:fig_3}
\end{figure}
\subsection{Overall Profit (Primary Metric)}

\textbf{Aggregate behavior.} The mean test profit with 95\% confidence intervals across seeds (\textbf{Fig. 1}) shows MAPPO+GAT outperforming MAPPO on average. The paired profit-difference distribution (GAT $-$ MAPPO) in \textbf{Fig. 2} is right-shifted, indicating a positive typical lift, although not universal.

\textbf{Seed-wise robustness.} \textbf{Fig. 3} depicts the difference per seed; 10/15 seeds favor GAT (\textbf{Fig. 4}), with several sizable positive deltas and a minority of negative cases. CRN pairing narrows the variance of differences, supporting a cleaner comparison across identically sampled episodes.

Conditioning policies on the co-purchase graph via attention yields directional gains in portfolio profit. The gains are consistent enough to be practically meaningful while acknowledging variability across seeds and start states.
 
\begin{figure}[htbp]
\centering
\includegraphics[height=0.4\textheight,keepaspectratio]{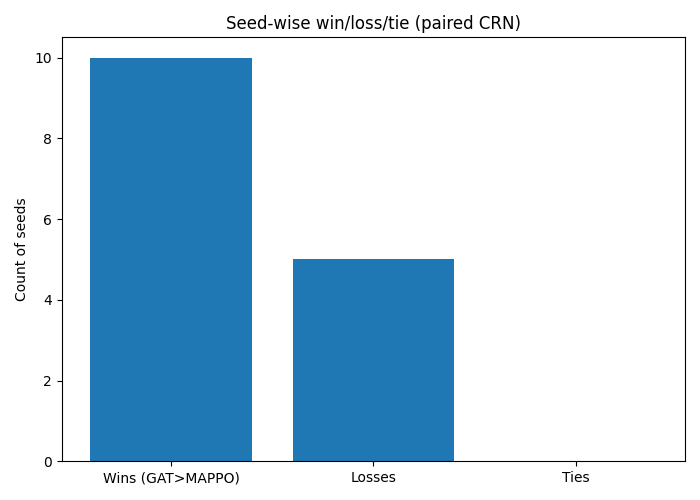}
\caption{Seed-wise win/loss/tie counts under CRN pairing (wins when GAT > MAPPO).}
\label{fig:fig_4}
\end{figure}

\begin{figure}[htbp]
\centering
\includegraphics[height=0.4\textheight,keepaspectratio]{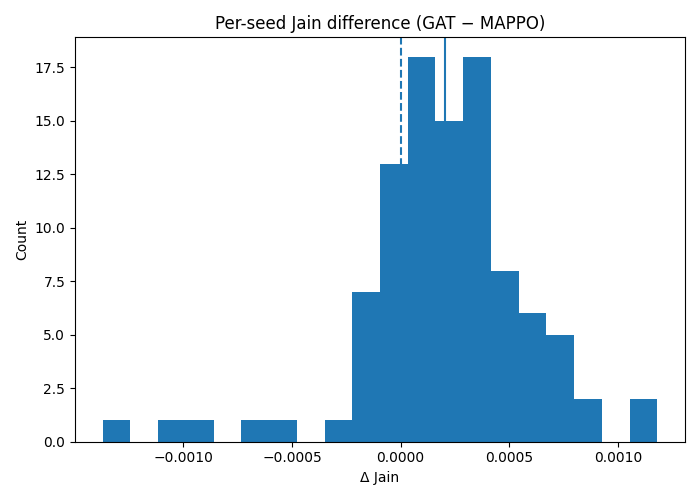}
\caption{Per-seed Jain stability (GAT $-$ MAPPO); positive values indicate improved fairness.}
\label{fig:fig_5}
\end{figure}

\begin{figure}[htbp]
    \centering
    \includegraphics[height=0.4\textheight,keepaspectratio]{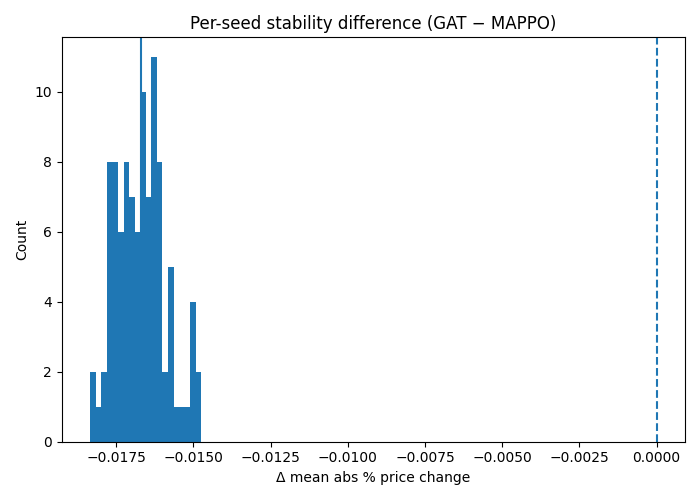}
    \caption{Per-seed stability (GAT $-$ MAPPO); negative values indicate smoother price paths.}
    \label{fig:fig_6}
    \end{figure}

\subsection{Fairness Across Products}
\textbf{Fig. 5} shows the Jain histogram (GAT $-$ MAPPO). The mass concentrates near zero with a slight positive shift, indicating no degradation in equity and a small tendency toward improved fairness under graph attention. In retail contexts, this implies profit is not achieved by systematically sacrificing a subset of SKUs.

\subsection{Price Stability}
\textbf{Fig. 6} plots stability (GAT $-$ MAPPO) where lower is better (less price motion). The distribution lies mostly below zero, showing that GAT typically reduces average percentage price changes. This is managerially desirable: smoother prices reduce sticker shock and operational churn while maintaining profit.

\subsection{Practical Takeaways}
\begin{itemize}
    \item Profit: GAT provides a reliable positive lift in average test profit over MAPPO in this setting.
    \item Fairness: Profit gains do not come at the expense of SKU-level equity (Jain's index maintains or slightly improves).
    \item Stability: Policies with GAT exhibit smoother price paths on average.
    \item Feasibility: A single GAT layer over a sparse top-k item graph yields these benefits without brittle tuning and is tractable for modest catalogs (tens of SKUs).
\end{itemize}
Across seeds, MAPPO+GAT shows (i) consistent stability improvements, (ii) no fairness degradation with a slight positive trend, and (iii) directionally higher profit with heterogeneous per-seed outcomes (10/15 wins; modest average lift). These results support the hypothesis that relational inductive bias stabilizes control and can enhance performance when product relations are informative.

\section{Conclusion}
This paper explored dynamic retail pricing using an explicit cross-product structure. It built on a strong multi-agent baseline (MAPPO) by adding a lightweight Graph Attention layer (MAPPO+GAT) over a co-purchase graph. In a real-data environment (UCI Online Retail II; 60 SKUs), we selected the best model based on validation steps and tested it using multiple seeds. The results showed that MAPPO+GAT provides a notable increase in mean test profit compared to MAPPO. It also maintained, and slightly improved, fairness across SKUs while decreasing average price fluctuations. These results suggest that managing prices at the portfolio level benefits from understanding interactions among products, rather than relying solely on individual product features. A single, sparse GAT layer can convert this relational insight into measurable operational improvements for small catalog sizes.

\textbf{Practical implications.} For small and mid-size retailers, the proposed graph-aware policy is feasible to train and deploy: (i) it builds on standard PPO infrastructure, (ii) adds minimal overhead via sparse top-k neighbors and a single GAT layer, and (iii) achieves smoother price paths that align with merchandising practice and customer experience goals. Because improvements persist across seeds under CRN pairing, the method is robust rather than reliant on narrow hyperparameter settings

\textbf{Limitations.} The gains are not universal across seeds; some runs favor the MAPPO baseline. The setting is single-retailer and demand is simulated from fitted models without stockouts, lead times, or competitor learning. The item graph is static and derived from co-purchase counts; alternative graphs (content, embeddings, substitutions) were not exhaustively explored. Finally, experiments focus on tens (not hundreds or thousands) of SKUs.

\subsection{Future work}
\begin{itemize}
\item Richer environments: integrate inventory dynamics, replenishment costs, and service-level constraints; model multi-retailer competition and market share.
\item Graph design and dynamics: compare co-purchase, substitution, and learned embedding graphs; study time-varying graphs and cold-start SKUs.
\item Architectures and credit assignment: evaluate deeper/wider attention, multi-head cross-SKU critics, or message-passing among agents; investigate counterfactual or Shapley-style attributions for interpretability.
\item Scalability: benchmark on larger catalogs with block-sparse attention and sampling to maintain training speed; quantify wall-clock trade-offs.
\item Robustness and generalization: stress-test under price shocks, holiday regimes, and covariate shift; apply domain randomization and distributionally robust objectives.
\item Operational metrics: incorporate customer-facing stability constraints, fairness regularizers, and business rules (price bounds, MAP compliance) directly into the objective.
\end{itemize}

Overall, the evidence supports graph-aware multi-agent reinforcement learning as a practical next step for dynamic retail pricing: it captures cross-product interactions that matter in practice, improves portfolio profit without sacrificing stability, and remains tractable in lightweight configurations suitable for real deployments.

\acks{The author thanks the contributors of open-source reinforcement learning frameworks and graph neural network toolkits that made this research possible. The work was carried out independently and did not receive any external funding or institutional support.}


\bibliography{GAT.bib}

\end{document}